\documentclass{article}
\usepackage{spconf,amsmath,graphicx}
\usepackage{mathtools}
\usepackage{comment}
\usepackage{enumitem}
\usepackage[flushleft]{threeparttable}
\usepackage{cite}
\usepackage{soul}
\usepackage{hyperref}       % hyperlinks
\usepackage{url}            % simple URL typesetting
\usepackage{booktabs}       % professional-quality tables
\usepackage{amsfonts}       % blackboard math symbols
\usepackage{nicefrac}       % compact symbols for 1/2, etc.
\usepackage{microtype}      % microtypography
\usepackage{xcolor}         % colors
\usepackage{graphicx}
\usepackage{amsmath}
\usepackage{multirow}
\usepackage[font=small,labelfont=bf]{caption}
\usepackage{subcaption}

\usepackage{newfloat}
\usepackage{floatrow}
\usepackage{float}

\newfloatcommand{capbtabbox}{table}[][\FBwidth]
\usepackage{cite}

\title{Robust Acoustic and Semantic Contextual Biasing in Neural Transducers for Speech Recognition}

\name{\begin{tabular}{c}Xuandi Fu$^{\dagger}$, Kanthashree Mysore Sathyendra$^{\dagger}$, Ankur Gandhe, Jing Liu, Grant P. Strimel, \\
Ross McGowan, Athanasios Mouchtaris \end{tabular}\thanks{$^{\dagger}$Equal Contribution}}

\address{Amazon Alexa}  
\begin{document}
\ninept
\maketitle
\begin{abstract}
\end{abstract}
Attention-based contextual biasing approaches have shown significant improvements in the recognition of generic and/or personal rare-words in End-to-End Automatic Speech Recognition (E2E ASR) systems like neural transducers. These approaches employ cross-attention to bias the model towards specific contextual entities injected as bias-phrases to the model. Prior approaches typically relied on subword encoders for encoding the bias phrases. However, subword tokenizations are coarse and fail to capture granular pronunciation information which is crucial for biasing based on acoustic similarity. In this work, we propose to use lightweight character representations to encode fine-grained pronunciation features to improve contextual biasing guided by acoustic similarity between the audio and the contextual entities (termed \textit{acoustic biasing}). We further integrate pretrained neural language model (NLM) based encoders to encode the utterance's semantic context along with contextual entities to perform biasing informed by the utterance's semantic context (termed \textit{semantic biasing}). Experiments using a Conformer Transducer model on the Librispeech dataset show a 4.62\% - 9.26\% relative WER improvement on different biasing list sizes over the baseline contextual model when incorporating our proposed acoustic and semantic biasing approach. On a large-scale in-house dataset, we observe 7.91\% relative WER improvement compared to our baseline model. On tail utterances, the improvements are even more pronounced with 36.80\% and 23.40\% relative WER improvements on Librispeech rare words and an in-house testset respectively.
\setlength{\textfloatsep}{10pt}


\section{Introduction}

% \section{Related Work}
% An example of semantic labels

E2E ASR systems are gaining popularity due to their monolithic nature and ease of training, making them promising candidates for deployment in commercial voice assistants (VAs). However, E2E models typically rely on word-piece vocabularies causing rare entities to often decompose into target sequences that are infrequent in training data making it difficult for the model to recognize them correctly ~\cite{le2021deep,  pundak2018deep, jain2020contextual, sainath2018no, bruguier2016learning}. 
% rare-phrases, proper nouns, named entities etc. correctly ~\cite{le2021deep,  pundak2018deep, jain2020contextual, sainath2018no, bruguier2016learning}. 
To provide the best user experience, it is important for a voice assistant to incorporate each user's custom environment and preferences, and use them to improve recognition of personalized requests like \textit{``call [Contact Name]"}, and \textit{``play my [Play List] on spotify"}. It is also desirable to recognize other types of rare-phrases like trending words (eg. ``coronavirus") or rare-words that appear on the screen (eg. rare-phrases from a wikipedia page displayed on the VA's screen or rare-phrases from a previous TTS response). Contextual biasing is widely used to inject such lexical contexts to bias the model's predictions towards them. Examples of lexical contexts include user defined terms (eg. user playlists, user's contacts) and generic rare-phrases (on-screen rare-words, trending words etc.). 

Neural contextual biasing methods for E2E ASR models broadly fall into two categories -- graph fusion approaches (eg. Deep Shallow fusion, Trie and WSFT-based approaches) \cite{zhao2019shallow, he2019streaming, gourav2021personalization} and fully-neural attention-based approaches \cite{pundak2018deep,jain2020contextual,chang2021context,sathyendra2022contextual,han2022improving,9747726}. In the first category, ~\cite{Zhao2019, le2021deep} introduced deep shallow fusion and trie-based contextual biasing with deep personalized LM fusion, respectively.
% Attention-based contextual biasing has become increasingly popular in recent years due to its fast adaption ability and improved recognition of rare words and contextual named entities. % 
More recently, fully-neural attention-based contextual biasing methods are becoming popular due to their improved rare-word recognition and ease of integration with E2E neural inference engines. Neural contextual biasing for LAS models was explored in~\cite{pundak2018deep, 44926} where bias phrases and/or contextual entities are encoded via BiLSTM encoders, and the model is made to bias toward them via a location-aware attention mechanism~\cite{chorowski2015attention, chen2019joint,bruguier2019phoebe}. For RNN-T models, ~\cite{jain2020contextual, chang2021context, sathyendra2022contextual} introduced fully-neural attention-based contextual biasing. ~\cite{chang2021context, sathyendra2022contextual} introduce contextual biasing for both the audio encoder and the text-based prediction network in transformer transducers (T-T) and conformer transducers (C-T). ~\cite{sathyendra2022contextual} explored contextual adaptation of pretrained RNN-T and C-T models using contextual adapters that are faster, cheaper and data-efficient.
% contextual adapters that are faster, cheaper to train and more data efficient through the use of a pre-trained neural transducer and an adaptation stage
%Attention-based biasing ~\cite{pundak2018deep, jain2020contextual, chang2021context, sathyendra2022contextual} typically consists of a sub-word context-encoder to encode contextual entities (eg. user's contact-names, favorite songs or generic bias-phrases etc.). An attention-based biasing layer uses model's intermediate representations – such as encoder, prediction network, or joint network outputs – as queries, and the encoded contextual entities as both keys and values. The biasing layer, through the attention mechanism, measures the correlation between the queries and the keys, to determine the contextual entities to attend over. %All prior approaches provide the same context embedding for keys and values for the biasing layers. Further, these approaches also share the context encoder used for biasing both the audio encoder and the prediction network.
% \label{tab:subword example}
% \input{tab_subword_example}

% Prior work on contextual biasing for neural transducers explored integrating attention-based contextual biasing adapters to learn the acoustic similarity between biasing entities and audio frames on the encoder side and semantic similarity with the decoded word-pieces (capturing the utterance's meaning or semantic context) on the prediction network side\cite{chang2021context,sathyendra2022contextual}. 
To embed the contextual entities, prior approaches relied on a text encoder (typically LSTM-based) that used grapheme-based subword/sentence-piece tokenizations\cite{pundak2018deep,jain2020contextual,sathyendra2022contextual,9747726}. However, two acoustically similar words may have completely different and non-overlapping subword tokenizations. For instance, the words ``seat" and ``meat", although acoustically similar, have very different subword tokenizations [sea, t] and [meat]. Assuming ``seat" is a new incoming OOV word, it may be incorrectly predicted as ``sit" (which could be seen more often in training data). A character-level encoder may capture the similarities in pronunciations in ``seat" and ``meat" better. Joint grapheme and phoneme context embedding has also been proposed by \cite{chen2019joint, bruguier2019phoebe} to incorporate pronunciation features. However, this requires a pronunciation lexicon or a grapheme-to-phoneme network. Further, a single word can have multiple pronunciations, and can result in longer lists of biasing-phrases. To overcome these issues, we use character-embedding models that not only capture fine-grained pronunciation information, but are also lightweight owing to their small character vocabularies, and require fewer model parameters. 

% In table \ref{tab:subword example}, ``sit" and ``meat" are context entities that are trained by the model while during evaluation a new context ``seat" are added to the biasing list. It's difficult for the model to rely on its subword tokens, [``sea", ``t"], to bias toward the correct words since the ``sea" is rarely trained, instead it will turn to predict a familiar homophone context ``sit".
Approaches have also been proposed to leverage semantic context features with language models to improve rare word recognition in ASR systems\cite{shenoy2021adapting, le2021contextualized}. In \cite{jain2020contextual, chang2021context, sathyendra2022contextual}, the transducer prediction network output is directly used as query in the prediction-network-side contextual biasing layer. However, the ability of the prediction network in encoding semantic context is limited by its training data, where the semantic context for long-tail utterances and rare words is not well trained \cite{ghodsi2020rnn}. This necessitates the integration of a pretrained language model(PLM) trained on ample text-only data that is capable of encoding semantics better. Using representations from the PLM as the query to the prediction-network-side biasing layer can help perform better semantic biasing. More generally, semantic biasing addresses the problem of ``given the previously decoded word-pieces, what contextual entity from the list is most likely to occur next?". It ensures semantic coherence between the word-pieces predicted by the model and the selected contextual entity. 

In this paper, we propose to train a character-based context representation on the encoder-side to capture generalizable acoustic context features useful for biasing based on acoustic similarity between the contextual entity and the incoming audio. We further introduce neural language model encoders to encode the utterance context and lexical contexts used for biasing based on semantics and meaning of the utterance, thus generalizing to open domain use cases.

\vspace{-1.5mm}
\section{Methodology}
\begin{figure}[t]
\centering
\includegraphics[width=0.99\linewidth]{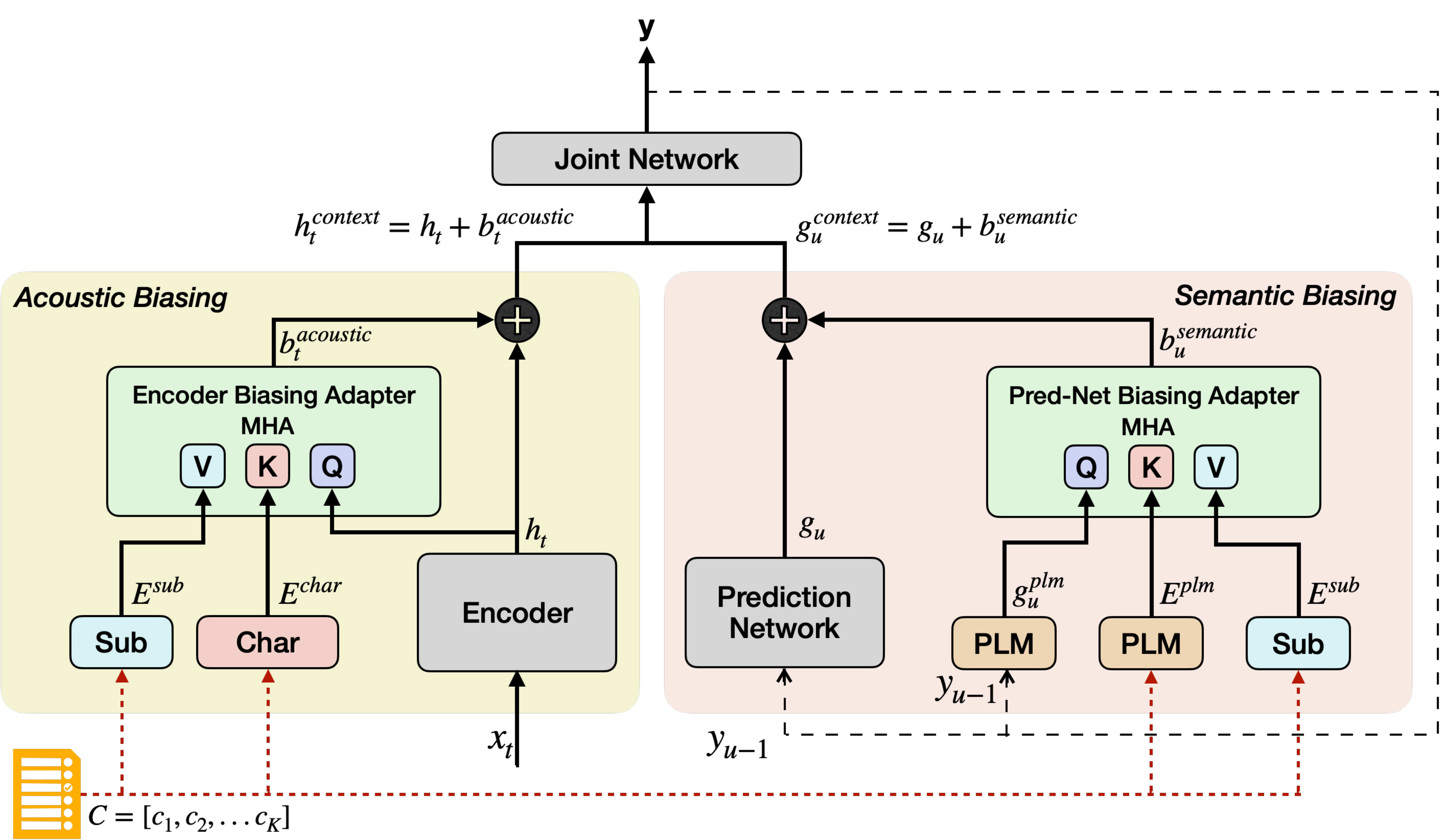}
\caption{The proposed acoustic and semantic contextual biasing model. \textit{Char}, \textit{Sub} and \textit{PLM} denote the character-, subword- and PLM-encoder respectively. Same color denotes shared parameters.}
\label{fig:high_level}
\vspace{-2.0mm}
\end{figure}

Neural transducer models typically consist of an encoder network ($H$), a prediction-network ($P$) and joint network ($J$) \cite{rnnt}. The inputs to the encoder networks are audio features of length T, $\boldsymbol{X}=(x_0, x_1,..,x_T)$. The encoder network $\emph{H}$ (typically stacked recurrent neural networks or conformer blocks), encodes the audio frame $x_t$ into D-dimensional intermediate representation $h_t = \emph{H}(x_t)$. The prediction-network encodes the sequence of previously predicted non-blank word-pieces $\boldsymbol{Y}=(y_0, y_1,..,y_{u-1})$ into hidden states $g_u$. The joint network fuses the encoder hidden states at time step $t$ and prediction-network hidden states at step $u$ to compute the probability of next word-piece as, $z_{t,u} = J(h_t + g_u)$. $J$ is a series of feed-forward layers with optional non-linear activations. The transducer model is optimized with the RNN-T loss that computes the alignment probability using the forward-backward algorithm \cite{rnnt}. In order to contextualize the transducer model and help recognition of contextual entities, a list of K bias-phrases, $\boldsymbol{C} = \{c_{1}, c_2, ..., c_K\}$, are encoded and injected to the transducer model ~\cite{sathyendra2022contextual, chang2021context}. The encoder-side biasing and prediction network-side biasing adapters are responsible for biasing towards the correct contextual entity by learning acoustic and semantic context-aware encoder and prediction network representations, $h_t^{context}$ and $g_u^{context}$ respectively, as shown in Figure \ref{fig:high_level}.
\vspace{-5pt}
\subsection{Acoustic Biasing}
% In this work, the impact of attention mechanisms on the contextual biasing models is into two components, where \emph{key} and \emph{query} only influence acoustic/semantic similarity computation whereas the \emph{value}, of which the attention-weighted sum is added to the encoder/prediction network outputs, influences the predictions of transducer model directly. 
The encoder biasing module relies on acoustic similarity between current audio features and the biasing list to attend towards the most acoustically similar contextual entity. The attention-based biasing adapter computes the cross-modal (from audio and text representations) similarity by using encoder state as the query, and the biasing list as keys and values. 
%The subword/sentence-piece encoder has limited ability to encode pronunciation features and to be linked with the utterance audio context. 
We use character-encoders to encode the bias-phrases and use it as the key for attention computation in the encoder biasing adapter, linking the acoustic hidden states $h_t$ and the external context. The proposed encoder contextual adapter consists of three components, a character-level key encoder, a subword-level value encoder and a cross-attention based biasing adapter for incorporating contextual information into the encoder hidden states.
\vspace{-2mm}
\subsubsection{Character Key Encoder}
The character key encoder encodes each contextual entity, $c_k$, into character context embeddings $e_k^{char}$. Each contextual entity, $c_k$, is first split into a sequence of characters $c_k^{char}$ 
% = [c_k^{char,1}, .., c_k^{char, I}]$
% $c_k = [c_k^0, c_k^1,..., c_k^n]$ 
and passed through an embedding layer followed by stacked Bi-LSTM layers, the final state of BiLSTM outputs is taken as the key embeddings, denoted as $E^{char} = [e_1^{char}, e_2^{char},..., e_K^{char}] \in \mathbb{R}^{K \times D^{c}}$ where $e_k^{char} = BiLSTM(Embedding(c_k^{char}))$ and $D^{c}$ is the embedding size for the character embeddings. 
%One advantage for using character-based encoder instead of phoneme encoder to capture pronunciation features is that it doesn't require phoneme labels available for the training data or additional training steps to learn a separate graphme-to-phoneme encoder. --> removed because we have mentioend this in intro.
%a char tokenzier that has vocabulary size, $V^{char}$ , $c_k^{char} = (c_k^1, c_k^2, ..., c_k^n)$, $c_k^i \in [0,V^{char})$
% \begin{equation}
%     e_i^{char} = BiLSTM(Embedding(c_i^{char}))
% \end{equation}\
\vspace{-2mm}
\subsubsection{Subword Value Encoder}
\label{sec:subword-value-encoder}
The subword encoder is used for generating the value for the attention-based biasing layer, which informs the model's subword predictions (hence uses sub-word embeddings). We use the same sentence-piece model as the C-T model to be consistent with the output unit of the transducer model\cite{kudo2018subword, kudo2018sentencepiece}. The tokenized subword sequences are passed through an embedding layer followed by stacked BiLSTM layers, and the final state of the BiLSTM is forwarded as the subword-level context embedding denoted as $E^{sub} = [e_1^{sub},e_2^{sub},...,e_K^{sub}]$ and each embedding is a $D^s$ dimensional vector.
\vspace{-2mm}
\subsubsection{Encoder Biasing Adapter (EBA) for Acoustic Biasing}
The EBA learns to bias toward relevant context entities by relying on the acoustic similarity between the audio and the biasing list\cite{sathyendra2022contextual}. The multi-head attention (MHA) module takes in audio/encoder hidden states $h_t$ as the \textit{query} and a list of character context embedding for bias-phrases $E^{char}$ as \textit{keys} to compute per-entity attention scores $a_t^k$ (Equation \ref{eq:attn}) for $t \in [0, T]$. The \textit{query} and \textit{keys} are first projected via $\boldsymbol{W}^k$ and $\boldsymbol{W}^q$ correspondingly. The attention score $a_t^k$, computed by the scaled dot product attention mechanism, measures the similarity between the audio features at time step $t$ and bias phrase $c_k$. The biasing vector $b_t^{acoustic}$, is computed by the attention-weighted sum of the projected \textit{value} embeddings (projected via $\boldsymbol{W}^v$) and added back to $h_t$ to obtain context-aware features $h_t^{context}$ (Equation~\ref{eq:bias_vector}). 
% The biasing adapter on the encoder side computes similarity between the acoustic representation at the current step and the context representations in the biasing list based on multi-head attention mechanism \cite{vaswani2017attention} and fuse the encoder output with the contextual output. 
%To fuse the character-based and subword representation in the biasing adapter, we experimented to use the concatenated character and subword $E^{joint} = [E^{char}; E^{sub}]$ and solely either subword embedding $E^{sub}$ and character embedding $E^{char}$ as the key for the multi-head attention to compute the acoustic similarity with audio output. For value of the attention, subword context embedding are used for maintaining consistency between the contextual output and the output word-piece vocabulary.\\
\setlength{\abovedisplayskip}{2mm}
\setlength{\belowdisplayskip}{0mm} 
\begin{align}
\label{eq:attn}
        & a_t^k = Softmax(\frac{\boldsymbol{W}^ke_k^{char}\boldsymbol{W}^qh_t}{\sqrt{d}})\\
\label{eq:bias_vector}
        & b_{t}^{acoustic} = \sum_k^Ka_t^k\boldsymbol{W}^ve_k^{sub}; \quad h_t^{context} = h_t + b_{t}^{acoustic}
        % & h_t^{context} = h_t + \underbrace{\sum_i^Ka_t^k\boldsymbol{W}^ve_k^{sub}}_{b_{u}^{acoustic}}
\end{align}
% We also explore alternative \textit{key} embedding representations by concatenating character and subword context embeddings as, $e_k = [e_k^{char};e_k^{sub}]$ resulting in each bias-phrase embedding with ($D^{s} + D^{c}$) dimensions.
\vspace{-4mm}
\subsection{Semantic Biasing}
One limitation of E2E models is that the decoder/prediction-network has to be trained on transcribed audio data which covers limited carrier phrases or semantic utterance contexts as compared to a large-scale text-only corpus. Since acoustic biasing learns similarity between context embedding and audio frames, it tends to give equal attention to homophones and cannot differentiate between homophones. In these cases, a strong indication from the utterance's semantic context can help the model select the right entity (among words with similar pronunciations) to bias towards. For this, we introduce a pretrained language model(PLM)-based query and key encoder for prediction network biasing. 
%Besides, with the introduction of the character-based context embedding on the encoder side, it increases the possibility for homophone context to be activated based on acoustic similarity. 
% For long-tail utterances, particularly, the prediction network does not encode semantic context well.
%and compared it with subword value context vector.
\vspace{-2mm}
\subsubsection{PLM-based Query-Key Encoder}
The PLM-based encoder (henceforth referred to as PLM-encoder) encodes the previously predicted word pieces $(y_0, y_1, ..., y_{u-1})$ to produce an utterance semantic context vector $g_{u}^{plm} \in \mathbb{R}^{1 \times D^{p}}$, where $D^{p}$ is PLM embedding size. The same PLM-encoder is also used to encode the bias-phrases, resulting in a shared embedding space between the \textit{queries} and the \textit{keys}. The bias phrases in $\boldsymbol{C}$ are first tokenized into subword units using the same sentencepiece model as the transducer, and encoded with the PLM to obtain $E^{plm} \in \mathbb{R}^{K \times D^{p}}$. 
\vspace{-5.5mm}
\subsubsection{Pred-Net Biasing Adapter (PNBA) for Semantic Biasing}
The PNBA adopts a similar scaled-dot-product attention as the encoder side (Equation~\ref{eq:attn}) to compute the \textit{semantic similarity} between the utterance context and the bias phrases. The \textit{query} is the PLM-generated semantic embedding $g_{u}^{plm}$ encoding the utterance semantic context, \textit{key} is the PLM-generated context embedding $E^{plm}$. For \textit{value}, we use the subword value embedding $E^{sub}$, shared with the encoder-side biasing adapter. The semantic prediction network biasing vector, $b_u^{semantic}$ is then computed similar to Equation~\ref{eq:bias_vector} for $u$, and fused with the prediction network hidden states as $g_u^{context} = g_u + b_u^{semantic}$.
% \begin{align}
% \label{eq:decoder_attn}
%         g_u^{context} = g_u + \sum_i^Ka_u^k\boldsymbol{W}^ve_k^{sub} 
% \end{align}
The generated context-aware hidden states, $h_t^{context}$ and $g_u^{context}$, are passed to the joint network to predict the probability for the next word piece, $z_{t,u}^{context} = J(h_t^{context} + g_u^{context})$. The biasing adapters are trained with the RNN-T loss, through the contextual-adaptation approach proposed by ~\cite{sathyendra2022contextual}. 
% The two biasing adapters (including the char-based, sub-word-based and PLM context-encoders) are trainable, while the rest of the network is initialized from pretrained weights and are kept frozen.
% \begin{align}
% \label{eq:decoder_attn}
%       z_{t,u}^{context} = J(h_t^{context} + g_u^{context})
% \end{align}
% ==================================== unused ==================================== 
%\textbf{Decoder biasing}
%For biasing on the decoder side, we used subword embedding as both key and value to learn the semantic similarity between the context entities and the word-piece hidden states. Importantly, the subword context learns the continuation among word pieces from the context entities,e.g.. The biasing context vector is computed similiar as the encoder side based on multi-head attention mechanism, $h_u^{context}$.

%& C^{key} = [E^{char}; E^{sub}];  C^{value} = [E^{sub}] \\
\vspace{-3pt}
\section{Experimental Results}
\begin{table*}[t]
\centering
\tabcolsep=0.1cm
\caption{\textbf{Librispeech.} Absolute WER for the models for K = \texttt{\{50, 100, 500, 1000\}}. Inputs to the Encoder biasing adapters in the form (Key, Value), and Pred-Net biasing adapters in the form (Query, Key, Value). \textit{`Sub', `Char', `PLM' and `PredNet'} are abbreviations to denote sub-word embeddings $E^{sub}$, character embeddings $E^{char}$, PLM embeddings $E^{plm}$ and the Prediction network states $g_u$ respectively.}
\vspace{-6pt}
\label{tab:librispeech}
\resizebox{0.72\linewidth}{!}{%
\begin{tabular}{@{}lclccccccccc@{}}
\toprule
\multirow{2}{*}{\textit{\textbf{Model Type}}} & \multirow{2}{*}{\textit{\textbf{\begin{tabular}[c]{@{}c@{}}Encoder Biasing\\ Adapter Inputs (K, V)\end{tabular}}}} &  & \multirow{2}{*}{\textit{\textbf{\begin{tabular}[c]{@{}c@{}}Pred-Net Biasing \\ Adapter Inputs (Q, K, V)\end{tabular}}}} & \multicolumn{4}{c}{\textit{\textbf{test-clean}}} & \multicolumn{4}{c}{\textit{\textbf{test-other}}}\\ \cmidrule(l){5-8} \cmidrule(l){9-12} 
                                              &                                                                                                                       &  &                                                                                                                        & \textit{K=50}  & \textit{K=100}      & \textit{K=500}                                                  & \textit{K=1000}                    & \textit{K=50}  & \textit{K=100}      & \textit{K=500}                                                  & \textit{K=1000}                        \\ \midrule
\textit{C-T\cite{conformer}}                             & \textit{-}                                                                                                        &  & \textit{-}                         & \multicolumn{4}{c}{6.08}                        &  \multicolumn{4}{c}{14.01}                  \\\midrule
\textit{Baseline (B1) ~\cite{sathyendra2022contextual}}                             & \textit{(Sub, Sub)}                                                                                                   &  & \textit{-}                         & 4.70     & 4.75  & 5.05  & 5.19   & 12.08                                                                                                            & 12.21                                                & 12.75                                               & 13.05                        \\ 
\textit{Char-I}                               & \textit{(Char, Char)}                                                                                                 &  & \textit{-}                                                                       & 4.98  & 5.06   & 5.30   & 5.43    & 12.67                                                               & 12.70                                                 & 13.20                                               & 13.55                        \\
% \textit{Char-II}                              & \textit{({[}Char; Sub{]}, Sub)}                                                                                       &  & \textit{NA}                                                                                                                & 4.51                         & 11.97                        & 4.8                          & 12.44                        & 5.05                         & 12.64                        \\
\textit{Char-II}                             & \textit{(Char, Sub)}                                                                                                  &  & \textit{-}                                                                                      & 4.56    & 4.63  & 4.82    & 5.08      & 11.80                                             & 12.06                                        & 12.51                                                & 12.89                        \\ \midrule
\textit{Char-Subword}                         & \textit{(Char, Sub)}                                                                                                  &  & \textit{(PredNet, Sub, Sub)}                                                          & 4.27   & 4.31   & 4.67  & 4.95      & 11.16                                                       & 11.34                                               & 12.21                                             & 12.64                        \\
\textit{Subword-PLM}                         & \textit{(Sub, Sub)}                                                                                                  &  & \textit{(PredNet, Sub, Sub)}                                                          & 4.22     & 4.32    & 4.63  & 4.95      & \textbf{11.01}                                                   & 11.22                                               & 12.11                                              & 12.72                        \\
\textit{Char-PLM}                             & \textit{(Char, Sub)}                                                                                                  &  & \textit{(PLM, PLM, Sub)}                                                                    & \textbf{4.10}    & \textbf{4.11}    & \textbf{4.52}  & \textbf{4.83}     & \textbf{11.01}                                                  & \textbf{11.21}                                                & \textbf{12.09}                                            & \textbf{12.60}                        \\ \bottomrule
\end{tabular}}%
\vspace{-2.5mm}
\end{table*}

\begin{table}[t]
\centering
\tabcolsep=0.1cm
\caption{\textbf{In-house datasets}. WERR on a variety of in-house testsets. The inference contextual entities (C) is one of \textit{rare-words (RW), proper-names (PN) or device-names (D). K=100.}}
\vspace{-6pt}
\label{tab:in-house}
\resizebox{0.98\linewidth}{!}{%
\begin{tabular}{@{}llcccccc@{}}
\toprule
\multicolumn{2}{l}{\multirow{2}{*}{\textit{\textbf{Model}}}} & \multicolumn{1}{l}{\textit{\textbf{General}}} & \textit{\textbf{RW}} & \textit{\textbf{ZSRW}} & \textit{\textbf{Knowledge}} & \multicolumn{1}{l}{\textit{\textbf{Proper Names}}} & \multicolumn{1}{l}{\textit{\textbf{Devices}}} \\ \cmidrule(l){3-8} 
\multicolumn{2}{l}{}                                         & \textit{C = (RW)}                             & \textit{C = (RW)}    & \textit{C = (RW)}      & \textit{C = (RW)}           & \textit{C = (PN)}                                  & \textit{C = (D)}                              \\ \midrule
\textit{Baseline (B1)}                   &                   & 0.00\%                                        & 0.00\%               & 0.00\%                 & 0.00\%                      & 0.00\%                                             & 0.00\%                                        \\ \midrule
\textit{Char-I}                          &                   & -4.11\%                                       & -9.54\%              & -17.45\%               & -5.18\%                     & -13.78\%                                            & +3.10\%                                        \\
\textit{Char-II}                         &                   & +0.91\%                                       & +1.19\%              & +5.52\%                & +2.38\%                     & +3.24\%                                             & +1.65\%                                        \\ \midrule
\textit{Char-Subword}                    &                   & +5.18\%                                       & +8.48\%              & +16.40\%               & +6.63\%                     & +12.24\%                                            & \textbf{+5.58\%}                                        \\
\textit{Subword-PLM}                    &                   & +7.00\%                                       & +12.58\%              & +15.80\%               & +9.02\%                     & +8.91\%                                            & +3.10\%                                        \\
\textit{Char-PLM}                        &                   & \textbf{+7.91\%}                              & \textbf{+14.97\%}    & \textbf{+23.40\%}      & \textbf{+11.40\%}           & \textbf{+13.13\%}                                  & +4.75\%                               \\ \bottomrule
\end{tabular}
}%
\vspace{-1mm}
\end{table}
% ----- with slot-datasets ----

% version 1
% \begin{table}[t]
% \centering
% \tabcolsep=0.1cm
% \caption{Evaluation results(WERR) on in-house testsets}
% \vspace{-6pt}
% \label{tab:in-house}
% \resizebox{0.9\linewidth}{!}{%
% \begin{tabular}{lcccc}
% \toprule
% Model & General (\%) & RW (\%) & ZSRW (\%) & Knowledge(\%) \\ \midrule
% Baseline (B1) & 0.00 & 0.00 & 0.00 &  0.00   \\ %\midrule
% Char-I &  1.37 & 1.47 & 6.34 & 2.07  \\ 
% Char-II &  1.37 & 1.47 & 6.34 & 2.07  \\ 
% Char-III &  0.91 & 1.19 & 5.62 & 2.38  \\ \midrule
% Char-Subword & 5.18 & 8.48 & 17.13 & 6.63   \\ 
% Char-PLM & 7.91 & 14.97 & 24.07 & 11.40 \\ 
% \bottomrule
% \end{tabular}}%
% \end{table}

% \label{tab:librispeech-rw}
\begin{table}[h]
\centering
\tabcolsep=0.1cm
% \caption{Evaluation results on zero-shot rare words in Librispeech (Reported metrics are in the
% format: WER (\%))}
% \caption{\textbf{Few-shot generalization} on Librispeech (\textit{test-clean}). WER for $n$-shot rare words where $n$=0, $\le$1, $\le$5, $\le$10, $\le$20, or $\le$100. \textit{K=100}. }\vspace{-8pt}
\caption{\textbf{Few-shot generalization} on Librispeech (\textit{test-clean}). WER for $n$-shot rare words; $n\le$\texttt{\{0, 1, 5, 10, 20, 100\}}. \textit{K=100}. }\vspace{-8pt}
% \caption{\textbf{Few-shot generalization} on Librispeech. WER for $n$-shot rare words in \textit{test-clean} where $n\le${0, 1, 5, 10, 20, or 100}. \textit{K=100}. }\vspace{-6pt}
\label{tab:librispeech-rw}
\resizebox{1.0\linewidth}{!}{%
\begin{tabular}{@{}lcccccccc@{}@{}}
\toprule
\multirow{2}{*}{\textit{\textbf{Model}}} & \multicolumn{6}{c}{\textit{\textbf{ZSR-WER}}}     & \multirow{2}{*}{\textit{\textbf{R-WER}}} & \multirow{2}{*}{\textit{\textbf{NR-WER}}}   \\ \cmidrule(l){2-7}
& \textit{\textbf{$0$-shot}} & \textit{\textbf{$\le$1-shot}} & \textit{\textbf{$\le$5-shot}} & \textit{\textbf{$\le$10-shot}} & \textit{\textbf{$\le$20-shot}} & \textit{\textbf{$\le$100-shot}} &  \\ \midrule
\textit{C-T}    &  113.6  &  103.9  &  85.4  &  70.0   &  54.5 &  27.7   &  24.6   & 3.8 \\ \midrule
\textit{Baseline (B1)}          &  91.2  &  81.6   &  60.1  &  46.2   &  33.5 &  15.2   &  13.1  & 3.9       \\
\textit{Char-II}                 &  85.5  &  75.2   &  55.2  &  42.1   &  30.4 &  13.7   &   12.0  & 3.8                \\ \midrule
% \textit{Subword-Subword\cite{sathyendra2022contextual}}         &  85.5  &  72.5   &  52.5  &  39.6   &  28.4 &  12.3   &  10.7      \\
\textit{Char-Subword}            &  79.2  &  67.7   &  48.8  &  37.0   &  26.0 &  11.5   &   10.0   & 3.7   \\
\textit{Subword-PLM}            &  81.9   & 67.8  & 47.9  & 36.0  & 25.5  & 11.0  & 9.6 &3.7 \\
\textit{Char-PLM}               & \textbf{71.0}   & \textbf{58.8}   & \textbf{42.3}   & \textbf{31.5}  & \textbf{22.0}   & \textbf{9.6}   & \textbf{8.3}        & \textbf{3.6}           \\ \midrule
\textit{WERR}      & \textbf{+22.1\%} & \textbf{+27.9\%}  & \textbf{+29.6\%}   & \textbf{+31.8\%} & \textbf{+34.3\%}  & \textbf{+36.8\%}   & \textbf{+36.6\%}   & \textbf{+7.7\%}                \\ \midrule
\textit{Word Count}              & \textit{331}     & \textit{461}           & \textit{814}             & \textit{1154} & \textit{1772}     & \textit{4844}  & \textit{5752} & \textit{46815}  \\ \bottomrule
\end{tabular}}%
\vspace{-2mm}
\end{table}

\begin{figure}[t]
\centering
\includegraphics[width=0.98\linewidth]{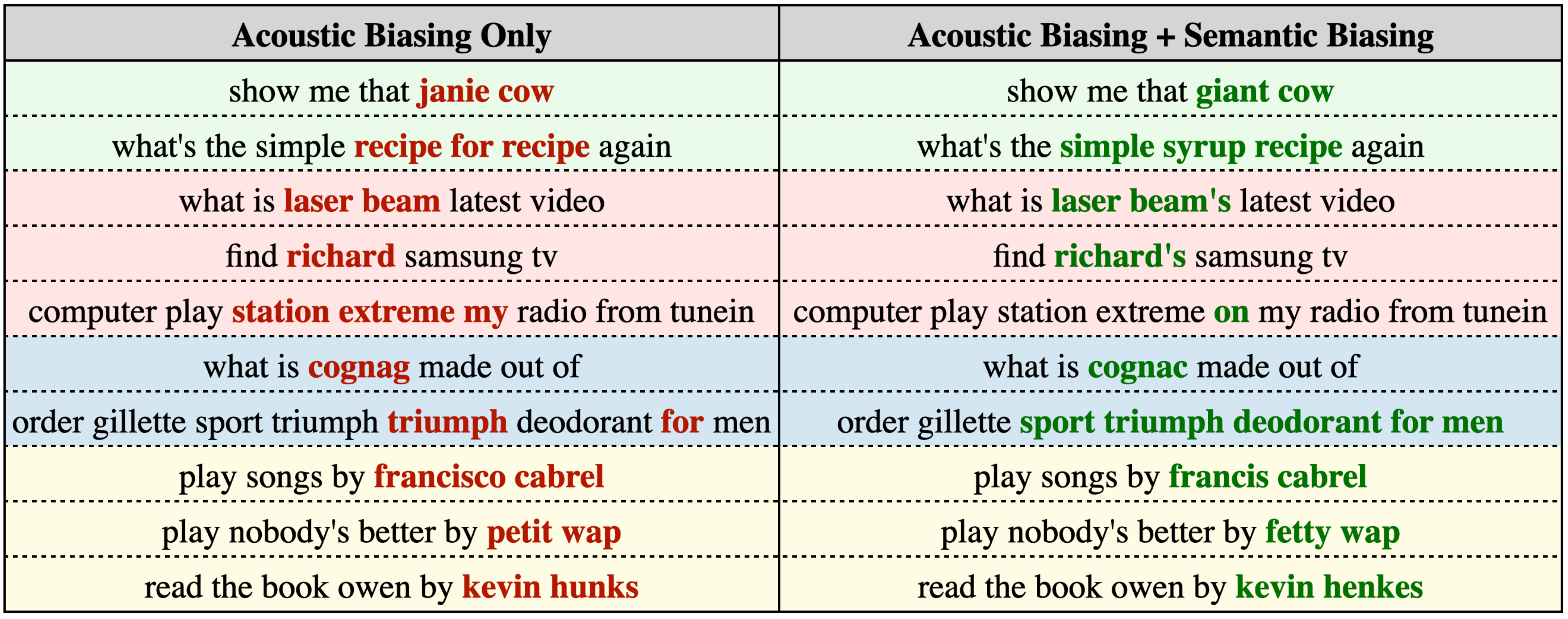}
\caption{Examples of semantic improvements (green), syntactic improvements (red), spelling corrections (blue) and external knowledge integration (yellow) from PLM Models.}
\label{fig:qual_ea}
\vspace{-2.5mm}
\end{figure}
\subsection{Datasets and Evaluation Metrics}
Our experiments are conducted on LibriSpeech \cite{panayotov2015librispeech} and an in-house large-scale voice assistant dataset. For the LibriSpeech datasets, a global rare word list is constructed by removing the top 5000 common words from the training vocabulary as in \cite{le2021contextualized}. During training, K = 50 randomly sampled distractors from the global rare words list are added to the biasing list, along with the correct words and the special $<$no-bias$>$ token \cite{pundak2018deep, sathyendra2022contextual}. The biasing list is formulated in a similar fashion during evaluation and includes the correct entity and $<$no-bias$>$. We further evaluate with larger biasing lists for K = \texttt{\{50, 100, 500, 1000\}}.\\
The in-house dataset consists of ~18k hours of de-identified audio data containing rare-words from multiple domains, such as \textit{song names}, \textit{location names}, \textit{device names}, \textit{playlist names} and other tail entities. The models are evaluated on four testsets with K = {100} biasing list size. \emph{General}: test utterances sampled follow the original training data distribution (contain both rare and non-rare utterances). \emph{Rare Word}: test utterances with at least one rare word (i.e. not in top-1000 most frequent words in training data). \emph{Zero-shot Rare Words}: testset containing tail utterances with rare words that are not in training data. \emph{Knowledge}: To test the semantic biasing performance of the models, we also prepare a test set with diverse semantic contexts, by pooling utterances from domains like \textit{Wikipedia}, \textit{Knowledge} and \textit{Books}. We also test the performance of our models when supplying personalized contexts on datasets -- \textit{Proper Names} and \textit{Devices}.\\
We evaluate the model performance on four metrics: (1) \textit{WER}: overall word error rate on the whole test set; (2) \textit{R-WER}: word error rate for rare words(RW) requiring biasing (3) \textit{NR-WER}: word error rate for non-rare words that do not require biasing (4) \textit{ZSR-WER}: word error rate for zero-shot rare words i.e. words not seen during training. For the in-house dataset, results are presented as relative error rate reductions\footnote{Due to internal company policies, we are unable to report absolute WER metrics on in-house data} (\textit{WERR, R-WERR, NR-WERR and ZSR-WERR})\footnote{Given a model A's WER ($\text{WER}_A$) and a baseline B's WER ($\text{WER}_B$), the WERR of A over B is computed as $\text{WERR} = (\text{WER}_B - \text{WER}_A)/\text{WER}_B$.}.
% \begin{table}[t]
% \centering
% \tabcolsep=0.1cm
% \caption{Left: Accuracy of the attention mechanism in selecting the correct entity for biasing, for different \textit{(Query, Key)} combinations in PNBA; Right: Results of combining proposed approach with SF}
% \vspace{-6pt}
% \label{tab:decoder-analysis}
% \begin{minipage}[b]{35mm}
% \resizebox{1.0\linewidth}{!}{%
% \begin{tabular}{@{}lcc@{}}
% \toprule
% \textit{\textbf{Models}}    & \textit{\textbf{AAI(\%)}} & \textit{\textbf{AASI(\%)}} \\ \midrule
% \textit{(PredNet, Sub)} & 0.00                                  & 0.00                                      \\
% \textit{(PredNet, PLM)} & +2.25                                 & +13.80                                  \\
% \textit{(PLM, PLM)}     & \textbf{+7.49}                                 & \textbf{+16.55}                                  \\ \bottomrule
% \end{tabular}}%
% \end{minipage}
% \begin{minipage}[b]{45mm}
% \resizebox{1.0\linewidth}{!}{%
% \begin{tabular}{@{}lcc@{}}
% \toprule
% \textit{\textbf{Models}}    & \textit{\textbf{AAI(\%)}} & \textit{\textbf{AASI(\%)}} \\ \midrule
% \textit{(PredNet, Sub)} & 0.0                                  & 0                                      \\
% \textit{(PredNet, PLM)} & 2.25                                 & 13.80                                  \\
% \textit{(PLM, PLM)}     & 7.49                                 & 16.55                                  \\ \bottomrule
% \end{tabular}}%
% \end{minipage}
% \vspace{-3mm}
% \end{table}

\begin{table}[t]
\centering
\tabcolsep=0.1cm
\caption{Accuracy of the attention mechanism in selecting the correct entity for biasing, for different Query-Key \textit{(Q-, K-)} pairs for PNBA.}
\vspace{-6pt}
\label{tab:decoder-analysis}
\resizebox{0.64\linewidth}{!}{%
\begin{tabular}{@{}lcc@{}}
\toprule
\textit{\textbf{Models}}    & \textit{\textbf{AA Relative(\%)}} & \textit{\textbf{AAS Relative(\%)}} \\ \midrule
\textit{(Q-PredNet, K-Sub)} & 0.00                                   & 0.00                                       \\
\textit{(Q-PredNet, K-PLM)} & +2.25                                 & +13.80                                  \\
\textit{(Q-PLM, K-PLM)}     & \textbf{+7.49}                                 & \textbf{+16.55}                                  \\ \bottomrule
\end{tabular}}%
\vspace{-2.5mm}
\end{table}

% \begin{table}[t]
% \centering
% \tabcolsep=0.1cm
% \caption{Attention metrics for prediction network biasing (Prediction network biasing model are defined as (Query, Key) and subword context embedding are used by all the following models.)}
% \vspace{-6pt}
% \label{tab:decoder-analysis}
% \resizebox{0.9\linewidth}{!}{%
% \begin{tabular}{lccc}
% \toprule
% Models & Attn. Acc Imp.(\%) & Attn. Score Imp.(\%)  \\ \midrule
% (Q-PredNet, K-Sub) & 0.0 & 0  \\ 
% (Q-PredNet, K-PLM) & 2.25 & 13.80 \\
% (Q-PLM, K-PLM) & 7.49 & 16.55 \\
% \bottomrule
% \end{tabular}}%
% \vspace{-2.5mm}
% \end{table}
\vspace{-5pt}
\subsection{Baseline and Model Configuration}
\vspace{-3pt}
The proposed methods are compared against the contextual biasing adapter baseline using only the subword encoder to encode contextual entities \cite{sathyendra2022contextual}. We follow the same adapter-stype training strategy as in \cite{sathyendra2022contextual} to first train the core transducer models and then train the adapters on \emph{Mixed} dataset with the base transducer parameters frozen to avoid degradation on recognizing general words.
\vspace{4pt}\\
\textbf{ASR model.} The input audio features are 64-dimensional LFBE features extracted every 10 ms with a window size of 25 ms. The features of each frame are then stacked with the left two frames, followed by a downsampling factor 3 to achieve a low frame rate, resulting in 192-dimentional features. For training the biasing layer, we used \emph{Mixed} dataset of utterances that contain rare words and general utterances containing common words to teach the model to learn the representations for context entities and the $<$no-bias$>$ token. The contextual biasing adapters are trained with streaming conformer-transducer (C-T) architecture to collaboratively decode the word-pieces. For LibriSpeech, we use 14 conformer blocks \cite{conformer} for the audio encoder. The hidden state size $d_{model}$, the projection dimension $d_{ff}$ and the number of attention heads $h$ for the conformer encoder are 256, 1024 and 4 correspondingly. For the prediction-network, the transcriptions are first tokenized by 2500 word-piece SentencePiece tokenzier~\cite{kudo2018sentencepiece}, and then passed through a 256-dim embedding layer followed by a 1-layer LSTM with 640 units and output size of 512-dim. For the in-house dataset, the audio encoder has 12 conformer blocks with $d_{model} = 512$, $d_{ff}=512$ and $h=4$, and output size is 512-dim. The prediction network encodes word-pieces with a 512-dim embedding layer and a 2-layer LSTM with 736 units and the output is projected to 512-dim.
\vspace{4pt}\\
\textbf{Char-Subword Encoder Biasing Model} For the char-subword biasing model for Librispeech, the context entities are tokenized with a 38-character tokenizer then embedded with a 64-dim embedding layer followed by one 64-unit BiLSTM layer. For subword representations, the contextual entities are tokenized using a 2500 sentence-piece tokenizer and encoded with a 64-dim embedding layer followed by 64-unit BiLSTM layer. The query, key and values for the multi-head attention layers in the acoustic biasing layer are projected to 128-dim tensors. For the in-house dataset, same-sized character-level context encoder and subword context encoder are used whereas the contextual entities are tokenized by a 4000 sentence-piece tokenizer for obtaining the subword representations. For training, we use the Adam optimizer with learning rate 5e-4 and 1e-4 for the LibriSpeech and the in-house model respectively.
\vspace{4pt}\\
\textbf{Pretrained Language Model (PLM) encoder:} For LibriSpeech, the PLM-encoder has a 256-dim embedding layer followed by a 2-layer LSTM (256 units each), trained on the text-only Wikitext-103 corpus\cite{merity2017pointer}. For in-house data, a domain-general NLM (embedding-dim 512, lstm-size 512) was trained on 80 million utterances with live audio from anonymized user interactions. To improve coverage on rare entities, we also added 25 million entities from artist and song name catalogs, and 8 million entities from place name catalogs.
\vspace{-5pt}
\subsection{Results}
Table \ref{tab:librispeech},\ref{tab:in-house} show WER for the Librispeech and in-house datasets. On Librispeech, the proposed acoustic-semantic biasing model outperforms the subword-embedding baseline~\cite{sathyendra2022contextual} for varying biasing list sizes(+9.26\% and +4.62\% WERR for test-clean K=100 and 1000). For the in-house dataset, it shows improvementments on both \emph{General} (+7.91\% WERR) and \emph{Rare-Word} testsets (+14.97\% WERR).
\\
%\\
% We especially compared the model's generalization ability on open-domain and zero-shot testset in \ref{tab:in-house} for in-house dataset and the ability for zero-shot rare words recognition in \ref{tab:librispeech-rw} for librispeech.\\
\textbf{Improved generalization abilities.} Character key encoder along with PLM-based prediction network biasing shows significant improvements over baseline \textit{(B1)} on recognizing zero-shot and few-shot rare words. On Librispeech, WER on 0\textit{-shot} rare words, improved from 91.2\% to 85.5\% for \textit{Char-II} when compared with B1, and further improved to 71.0\% with PLM-based prediction network biasing (Table~\ref{tab:librispeech-rw}). The relative improvement becomes more pronounced as training samples for the rare words increase (+36.8\% WERR for $\le$100\textit{-shot}). On the in-house ZSRW dataset, the \textit{Char-II} resulted in +5.52\% WERR. To tease out the improvements from PLM integration, we compared \textit{Char-PLM} (PLM-based query-key) with \textit{Char-Subword} model ($g_u$ as query, \textit{Sub} for key). On ZSRW testset, WERR goes from 16.40\% (for \textit{Char-Subword}) to 23.40\% for \textit{Char-PLM}(Table~\ref{tab:in-house}).\vspace{2pt}\\
\textbf{Importance of PLM-based Prediction Network biasing.} The PLM-based query-key biasing layers demonstrate improved recognition in terms of semantic, syntactic, spelling improvements and external knowledge integration. Some qualitative examples are shown in Figure~\ref{fig:qual_ea}. With the external knowledge learned from larger text corpus, the PLM-based biasing layers are able to correctly attend over the artist name ``\textit{francis cabrel}". On the \textit{Knowledge} in-house testset containing tail/rare words from multiple domains, the PLM-based model has shows 11.40\% WERR compared over B1.\vspace{2pt}\\
\textbf{Evaluation of attention mechanism.} 
To compare the ability of the proposed semantic biasing model to attend over the correct bias phrase (vs. over distractors) based on semantic similarity, we compute two additional metrics -- 1) \emph{Attention Accuracy (AA)(\%)}: \% of utterances where the model attends over the correct contextual entity, i.e. has the highest attention score for the correct entity. 
%ratio of the correct biasing phrase has highest attention score in the biasing list; 
2)\emph{ Avg. Attention Score (AAS)(\%)}: the average attention score for the correct bias phrase assigned by the attention models. Comparisons are made on the in-house \textit{ZSRW} testset (Table \ref{tab:decoder-analysis} - reported as relative improvement). The PLM-based biasing model(\textit{(Q-PLM, K-PLM)}) shows 7.49\% relative and 16.55\% relative improvement in \textit{AA} and \textit{AAS} respectively, when compared with using prediction-network output as query. Thus, demonstrating its ability to leverage the utterance semantic context in selecting the entities to attend over.
\vspace{-5.1pt}
\section{Conclusion}
In this work, we propose to improve conformer-transducer biasing models with character-based acoustic biasing and PLM-based semantic biasing model to link the utterance context with the correct biasing entities. Through experiments on LibriSpeech and large-scale in-house datasets, we demonstrated that the proposed acoustic-semantic biasing model can encode better context features and learns a superior biasing model that is guided by acoustic and semantic similarity. It also shows significant improvements when generalizing to tail rare words and open-domain use cases compared to the baseline.

\vfill\pagebreak

\bibliographystyle{IEEEbib}
\bibliography{reference.bib}

\end{document}